\documentclass[letterpaper, 10 pt, conference]{ieeeconf}  

\IEEEoverridecommandlockouts                              

\usepackage[english]{babel}
\usepackage{times} 
\usepackage{cite}
\usepackage{hyperref}
\usepackage[font=footnotesize]{subcaption}
\usepackage{calrsfs}
\DeclareMathAlphabet{\pazocal}{OMS}{zplm}{m}{n}

\usepackage{graphicx} 
\usepackage{epsfig} 
\usepackage{pgfplots}

\pgfplotsset{compat=1.15}
\graphicspath{{./figure/}}

\makeatletter
\let\MYcaption\@makecaption
\makeatother

\makeatletter
\let\@makecaption\MYcaption
\makeatother

\usepackage{mathptmx} 
\usepackage{amsmath} 
\usepackage{amssymb}  
\usepackage{nicefrac}
\usepackage{mathtools}
\usepackage{optidef}

\def\x#1{\texttt{\expandafter\string\csname#1\endcsname}&\expandafter$\csname#1\endcsname$}

\DeclareMathOperator*{\maximize}{maximize}

\usepackage{tikz}
\usetikzlibrary{quotes,angles,calc}

\usepackage{siunitx}
\usepackage{todonotes}

\usepackage{algorithm}
\usepackage[noend]{algpseudocode}

\makeatletter
\def\BState{\State\hskip-\ALG@thistlm}
\makeatother

\title{ \LARGE \bf
	A framework for power line inspection tasks with multi-robot systems from signal temporal logic 
	specifications
}

\author{Giuseppe Silano$^{1,3}$, Davide Liuzza$^2$, Luigi Iannelli$^3$, and Martin Saska$^1$%
	\thanks{$^1$Giuseppe Silano and Martin Saska are with the  Faculty of Electrical Engineering,	
	Czech Technical University in Prague, Czech Republic, email: {\tt\small \{giuseppe.silano, 
	martin.saska\}@fel.cvut.cz.}}
	\thanks{$^2$Davide Liuzza is with the ENEA Fusion and Nuclear Safety Department, Italy, email: 
	{\tt\small davide.liuzza@enea.it.}}
	\thanks{$^3$Giuseppe Silano and Luigi Iannelli are with the Department of Engineering, 
	University of Sannio, Italy, email: {\tt\small \{giuseppe.silano, 
	luigi.iannelli\}@unisannio.it.}}
	\thanks{This work was partially funded by the European Union's Horizon 2020 research and 
	innovation programme AERIAL-CORE under grant agreement no. 871479, and by ECSEL Joint 
	Undertaking research and innovation programme COMP4DRONES under grant agreement no. 826610.}
}

\begin{document}
	
	\maketitle
	\thispagestyle{empty}
	\pagestyle{empty}
	
	\begin{abstract}
		
		Inspection of power line infrastructures must be periodically conducted by electric 
		companies in order to ensure reliable electric power distribution. Research efforts are 
		focused on automating the power line inspection process by looking for strategies that 
		satisfy different requirements expressed in terms of potential damage and faults detection. 
		This problem comes up with the need of safe planning and control techniques for autonomous 
		robots to perform visual inspection tasks. Such an application becomes even more 
		interesting and of critical importance when considering a multi-robot extension.
		
		In this paper, we propose to compute feasible and constrained trajectories for a fleet of 
		quad-rotors leveraging on Signal Temporal Logic (STL) specifications. The planner allows to 
		formulate rather complex missions avoiding obstacles and forbidden areas along the path. 
		Simulations results achieved in MATLAB show the effectiveness of the proposed approach 
		leading the way to experimental tests on the hardware.
	
	\end{abstract}

	\medskip
	
	Electric companies invest significantly on the inspection and preemptive maintenance of the 
	power line infrastructure. The most common strategy is to perform aerial inspection of the 
	power line corridor, at regular intervals. The traditional (and the most common) approach to 
	inspection uses a manned helicopter, equipped with multiple sensors, e.g., light and radar 
	sensors, visual, infrared and ultra-violet cameras, mounted on gyroscope stabilized gimbals, 
	and an expert crew, for recording and documenting the relevant data captured from these sensors.
	This data is later manually examined to detect potential faults and damage on different power 
	line components (e.g., cables, towers, insulators). This process is not only extremely time 
	consuming and dangerous for human operators, but also very expensive (\$1,500 for one hour 
	flight) and prone to human error~\cite{Baik2018JIRS}.
	
	In the last two decades, multiple complementary research directions have been explored for 
	automating the task of visual inspection. One key direction has been on developing Unmanned 
	Aerial Vehicles (UAVs) capable of inspecting the power line corridor and the 
	towers~\cite{Baik2018JIRS}. UAVs can execute two different levels of inspections, i.e., 
	macro and micro, depending on the wing type. For example, a fixed-wing UAV cannot hover 
	itself, thus it is used to provide a macro-level inspection which can be surveying power lines 
	in a large area. On the other hand, a rotary-wing UAV can perform a micro-level inspection 
	such as checking for mechanical failures on transmission components. However, using UAVs to 
	achieve these tasks is particularly challenging due to their limited battery capacity 
	and to the fact that operating close to lines may be impractical for the strong electromagnetic 
	field and the several obstacles (e.g., branches, markers ball)~\cite{Baik2018JIRS}.
	
	For all such reasons, having UAVs capable of interpreting high-level, possibly vague, tasks 
	specifications, and nevertheless plan and execute appropriate actions for a particular context 
	in which the system is operating, is more and more important. Symbolic control proposes to 
	fulfill this need by automatically designing feedback controllers that lead to the 
	satisfaction of formal logic specifications. Temporal-logic (TL) based motion planning 
	techniques can be used for this purpose. In particular, Signal Temporal Logic (STL) provides 
	formal high-level languages that can describe planning objectives more complex than well-suited 
	point-to-point navigation algorithms~\cite{Karaman2011JRS}. The task specification is given as 
	a temporal logic formula w.r.t. the discretized abstraction of the robot motion modelled as a 
	finite transition system~\cite{Kloetzer2010TRO}. Then, a high-level	discrete plan is found 
	by off-the-shelf model-checking algorithms~\cite{Bhatia2011RAM} given the finite transition 
	system and the task specification. This discrete plan is then implemented through a 
	corresponding low-level hybrid controller.
	
	In this paper, we propose a framework to encode missions for a fleet of quad-rotors as STL 
	specifications. Then, using motion primitives for quad-rotor trajectory 
	generation~\cite{Mueller2015TRO}, we construct an optimization problem to generate 
	optimal strategies that satisfy such specifications. The proposed approach generates feasible 
	dynamic trajectories accounting for velocity and accelerations constraints of the vehicles that 
	satisfy missions limitations, too. Simulations results in MATLAB show the effectiveness of the 
	proposed approach.
	
	Let us consider a continuous-time dynamical system $\pazocal{H}$ and its discretized time 
	version $x^+=f(x,u)$, where $x, x^+ \in X \subset \mathbb{R}^n$ are the current and next state 
	of the system, respectively, $u \in U \subset \mathbb{R}^m$ is the control input and $f \colon 
	X \times U \rightarrow X$ is differentiable in both arguments. The system's initial state is 
	denoted by $x_0$ and takes values from some initial set $X_0 \subset \mathbb{R}^n$. Let $T_s 
	= t_{k+1} - t_k$ for all $k \in \mathbb{N}_{\geq 0}$ be the sampling period and $T \in 
	\mathbb{R}_{\geq 0}$ be a trajectory duration, we can write the time interval as $t = (0, T_s, 
	2T_s, \dots, NT_s)$ with $NT_s=T$. Therefore, given an initial state $x_0$ and a finite control 
	input sequence $\mathbf{u}=(u_0, u_1, \dots, u_{N-1})$, a trajectory of the system is the 
	unique sequence of states $\mathbf{x} = (x_0, x_1, \dots, x_N)$ with $x_{k+1}=f(x_k, u_k)$. 
	
	The trajectory generator is designed to make the closed loop system satisfying a specification 
	expressed in STL. STL is a logic that allows the succinct and unambiguous specification of a 
	wide variety of desired system behaviors over time, such as ``The quad-rotor reaches the goal 
	within 10 time units while always avoiding obstacles''. Formally, let $M = \{\mu_1, \mu_2, 
	\dots, \mu_L\}$ be a set of real-valued functions of the state $\mu_k \colon X \rightarrow 
	\mathbb{R}$. For each $\mu_k$ define the predicate $p_k \coloneqq \mu_k(x) \geq 0$. 
	Set $AP \coloneqq \{p_1, p_2, \dots, p_L\}$. Thus each predicate defines a set over the system 
	state space, namely $p_k$ defines $\{x \in X | \mu_k(x) \geq 0\}$. Let $I \subset \mathbb{R}$ 
	denote a nonsingleton interval, $\top$ the Boolean True, $p$ a predicate, $\neg$ and $\wedge$ 
	the Boolean negation and AND operators, respectively, and $\pazocal{U}$ the Until temporal 
	operator. An STL formula $\varphi$ is built recursively from the predicates using the grammar 
	$\varphi \coloneqq \top | p | \neg \varphi | \varphi_1 \wedge \varphi_2 | \varphi_1 
	\pazocal{U}_I \varphi_2$, where $p$ is a predicate and $\varphi_1$ and $\varphi_2$ are STL 
	formulas. Informally, $\varphi_1 \pazocal{U}_I \varphi_2$ means that $\varphi_2$ must hold at 
	some point in $I$, and until then, $\varphi_1$ must hold without interruption. The disjunction 
	($\vee$), implication ($\Longrightarrow$), Always ($\square$) and Eventually ($\lozenge$) 
	operators can be also defined. Formally, the \textit{pointwise semantics} of an STL formula 
	$\varphi$ defines what it means for a system trajectory $\mathbf{x}$ to satisfy 
	$\varphi$~\cite{Raman2014CDC}.
	
	Designing a controller that satisfies a STL formula $\varphi$ is not always enough. In a 
	dynamic environment, where the system must react to new unforeseen events, it is useful to have 
	a margin of maneuverability. That is, it is useful to control the system such that to maximize 
	a degree of satisfaction of the formula. The latter can be defined and computed using the 
	\textit{robust semantics} of TL~\cite{Fainekos2009TCS, Donze2010FMATS}.
	
	Starting from the definition of a robust STL formula, indicated as $\rho_\varphi(\mathbf{x}, 
	t)$, we can compute control inputs $\mathbf{u}$ by maximizing the robustness over the set of 
	finite state sequences $\mathbf{x}$. The obtained sequence $\mathbf{u}^\star$ is valid if 
	$\rho_\varphi(\mathbf{x}^\star, t)$ is positive, where $\mathbf{x}^\star$ and 
	$\mathbf{u}^\star$ obey to the dynamical system $\pazocal{H}$. Thus, the goal is to find a 
	provably correct control scheme for a robot system which makes it meeting a control objective 
	$\varphi$ expressed in temporal logic. So, let $\epsilon > 0$ be a desired minimum robustness, 
	we solve the following problem
	\begin{equation}\label{eq:optimizationProblem}
	\begin{split}
	& \maximize_{\mathbf{u} \in U^{N-1}} \;\; {\rho_\varphi(\mathbf{x})} \\
	&\quad \text{s.t.}~\qquad x_{k+1}=f(x_k, u_k), \forall k=\{0,1, \dots, N-1\}, \\
	&\qquad\;\;\qquad x_k \in X, u_k \in U, \forall k=\{0,1, \dots, N \}, \\
	&\qquad\;\;\qquad \rho_\varphi(\mathbf{x}) \geq \epsilon  
	\end{split}.
	\end{equation}
	The same formalization applies if $f$, $x$ and $u$ represent the overall dynamics, state and 
	input of a multi-robot system. Because $\rho_\varphi$ uses the non-differentiable functions 
	$\max$ and $\min$~\cite{Fainekos2009TCS, Donze2010FMATS}, 
	solving~\eqref{eq:optimizationProblem} can be achieved by using Mixed-Integer Programming 
	solvers, non-smooth optimizers or stochastic heuristics~\cite{Raman2014CDC}. However, as 
	recently shown in~\cite{Pant2017CCTA}, it is more efficient and reliable to approximate the 
	non-differentiable objective $\rho_\varphi$ by a smooth (infinitely differentiable) function 
	$\tilde{\rho}_\varphi$ and to solve the resulting optimization problem using Sequence Quadratic 
	Programming.
	
	To come up with a trajectory that satisfies the vehicle constraints, the motion	primitives 
	defined in~\cite{Mueller2015TRO} were considered. Such a method allows to obtain a rapid 
	generation and feasibility verification of motion primitives for quad-rotors. The results of 
	the modified optimization problem is a set of splines~\cite[eq.~(22)]{Mueller2015TRO} whose 
	parameters can be tuned to achieve a desired motion fixing a combination of position, 
	velocity, and acceleration at the start and end points. Therefore, the objective function can 
	be reformulated replacing $\rho_\varphi(\mathbf{x})$ with its smooth version accounting for the 
	mathematical formulation of the trajectory generator $L$, i.e., $\tilde{\rho}_\varphi 
	\left(L(\mathbf{x})\right)$. To track the obtained references, any of the quad-rotor control 
	algorithms within the literature~\cite{Nascimento2019ACR} can be used.
	
	We show the effectiveness of the proposed approach through two numerical examples: (i) a 
	multi-drone reach and avoid problem in a constrained environment (eq.~\eqref{eq:reachAvoid}, 
	{\url{https://youtu.be/SHThWKYw1v8}.}), and (ii) a multi-mission example where several drones 
	have to fly one of two missions in the same environment (see eq.~\eqref{eq:multidroneMission}, 
	{\url{https://youtu.be/uqYq5xmCckY}.}). 
	\begin{subequations}
		\begin{equation}\label{eq:reachAvoid}
		\resizebox{0.80\hsize}{!}{$
		\begin{split}
		&\varphi_\mathrm{safe}^{i,j} = \square_{[0, T]} \varphi_\mathrm{dist}^{i,j} = \square_{[0, 
		T]} \left(\lVert {^W\mathbf{r}}^i - {^W\mathbf{r}}^j \rVert  \geq \delta_\mathrm{min} 
		\right) , \\
		&\varphi_\mathrm{mra} = \wedge_{k=1}^N \varphi_\mathrm{ra}^k \bigwedge \wedge_{k=1}^N \left(
		\wedge_{i \neq j} \varphi_\mathrm{safe}^{i,j} \right) ,
		\end{split}
		$}
		\end{equation}    
		\begin{equation}\label{eq:multidroneMission}
		\resizebox{0.95\hsize}{!}{$
		\begin{split}
		\varphi_\mathrm{pli}^N &= \wedge_{k=1}^N \Bigl( \square_{[0,T]} \left(
			\varphi_\mathrm{dist}^k \wedge \varphi_\mathrm{ws}^k \right) \Bigr) \bigwedge 
			\lozenge_{[0,T]} \biggl( \left( \wedge_{k=1}^{\nicefrac{N}{2}} 
			\varphi_\mathrm{pole1}^k \wedge \varphi_\mathrm{pole4}^k \right) \\
			&  \bigwedge \left( \wedge_{k=\nicefrac{N}{2}+1}^N	\varphi_\mathrm{pole2}^k \wedge 
			\varphi_\mathrm{pole3}^k \right) \pazocal{U}_{[0,\nicefrac{T}{2}]} \left( 
			\wedge_{k=\nicefrac{N}{2}+1}^N	\varphi_\mathrm{pole3}^k \wedge 
			\varphi_\mathrm{pole2}^k \right) \biggr) \\ 	
		\end{split}.
		$}
		\end{equation}
	\end{subequations}
	
	\bibliographystyle{IEEEtran}
	\bibliography{bib.bib}

\end{document}